# An overlapping-free leaf segmentation method for plant point clouds

**Dawei Li**[1,2], **Yan Cao**[1], **Guoliang Shi**[1], **Xin Cai**[1,2*], **Yang Chen**[1], **Sifan Wang**[1], and **Siyuan Yan**[1]

[1] College of Information Sciences and Technology, Donghua University, Shanghai, 201620 China; daweili@dhu.edu.cn

[2] Engineering Research Center of Digitized Textile & Fashion Technology, Ministry of Education, Donghua University, Shanghai, 201620 China; daweili@dhu.edu.cn

[*] Correspondence: Xin Cai (e-mail: xcai@dhu.edu.cn);

**Abstract:** Automatic leaf segmentation, as well as identification and classification methods that built upon it, are able to provide immediate monitoring for plant growth status to guarantee the output. Although 3D plant point clouds contain abundant phenotypic features, plant leaves are usually distributed in clusters and are sometimes seriously overlapped in the canopy. Therefore, it is still a big challenge to automatically segment each individual leaf from a highly crowded plant canopy in 3D for plant phenotyping purposes. In this work, we propose an overlapping-free individual leaf segmentation method for plant point clouds using the 3D filtering and facet region growing. In order to separate leaves with different overlapping situations, we develop a new 3D joint filtering operator, which integrates a Radius-based Outlier Filter (RBOF) and a Surface Boundary Filter (SBF) to help to separate occluded leaves. By introducing the facet over-segmentation and facet-based region growing, the noise in segmentation is suppressed and labeled leaf centers can expand to their whole leaves, respectively. Our method can work on point clouds generated from three types of 3D imaging platforms, and also suitable for different kinds of plant species. In experiments, it obtains a point-level cover rate of 97% for *Epipremnum aureum*, 99% for *Monstera deliciosa*, 99% for *Calathea makoyana*, and 87% for *Hedera nepalensis* sample plants. At the leaf level, our method reaches an average Recall at 100.00%, a Precision at 99.33%, and an average F-measure at 99.66%, respectively. The proposed method can also facilitate the automatic traits estimation of each single leaf (such as the leaf area, length, and width), which has potential to become a highly effective tool for plant research and agricultural engineering.

## 1. Introduction

The plant phenotype is determined or influenced by both genes and environmental factors, and it reflects all physical, physiological and biochemical characteristics, and traits of the plant structure, composition, growth, and development process [1]. With the research goes deeper in functional plant genomics and molecular-level plant breeding, traditional phenotyping tools have become a main bottleneck in further improvements [2]. High-throughput plant phenotyping technology has becoming an effective measure to address the problem. The research of plant phenotyping is to comprehensively assess the complex traits of plants and intuitively measure the parameters of those traits [3]. For most kinds of plants, leaves make up the majority of the surface morphology and also form the main structure of the plant, and the observation toward leaves can easily unveil the growth status of crops [4]. Leaf characteristics including morphology, texture, and color, often implies biotic stress (plant disease and pests) or abiotic stress (drought) that affect plant growth. Therefore, automatic leaf segmentation, as well as identification and classification that built upon it, are able to provide immediate monitoring for plant growth status to guarantee the output. Since the early 1990s, researchers have been using various imaging tools to monitor and analyze plant growth [5], and the automatic individual leaf segmentation and classification methods based on 2D images become both the hotspot and difficulty at the same time for the community. The complicated spatial structures of plants put forward a high requirement for a leaf segmentation algorithm to work well. Moreover, none of the segmentation methods is suitable to all kinds of plants due to the vast diversity of plant species. Therefore, the design of a universal, accurate, and efficient leaf segmentation algorithm remains to be a big challenge.

Categorized by types of the data, the plant leaf segmentation methods are generally divided into the class that is based on 2D images and the other based on 3D point clouds. Methods that based on 2D images often leverage classic image processing, machine learning, and pattern recognition techniques to carry out individual leaf segmentation for several kinds of standard crops with simple structures (e.g., *Arabidopsis*, tobacco, and wheat). Pape *et al.* [6] used a large dataset to train a classifier for distinguishing the leaf boundaries, which decreased the segmentation error caused by occlusion among leaves and achieved satisfactory results for *Arabidopsis*. Viaud *et al.* [7] carried out leaf pre-segmentation for *Arabidopsis* using the watershed algorithm, and then improved the pre-segmentation result based on an ellipse model. Plant phenotyping software tools based on 2D images such as PlantCV v2 [8] and Leaf-GP [9] can perform individual leaf segmentation for *Arabidopsis* and wheat plants, and also can analyze their growth status based on the segmentation results. Dobrescu *et al.* [10] proposed a leaf counting method based on deep learning and achieved satisfactory results on *Arabidopsis* and tobacco plants. Yin *et al.* carried out multi-blade segmentation, alignment and tracking for video sequences of growing *Arabidopsis* samples; their method solves the occlusion problem of leaves in most cases for rosette-like plants [11]. However, methods based on 2D images are vulnerable to the complicated environmental factors and they usually restrict the imaging to a narrow range of angle. In addition, if large areas of leaf overlapping appear in the image, 2D methods are prone to segmentation failures. Thus, this type of segmentation methods only reports satisfactory results on several rosette plants and standard plants.

3D point clouds avoid the lack of depth information, which gives the ability to intrinsically resolve the issue of occlusion among leaves. Paproki *et al.* [12] used multi-view 3D reconstruction and 3D meshes to generate a point cloud of *Gossypium hirsutum*. They then segmented leaves by applying region growing, and separated the petioles and the stems by using tubular fitting. Li *et al.* [4] first applied facet over-segmentation [13] on plant point clouds, and realized an individual leaf segmentation algorithm with facet region growing for greenhouse ornamentals (e.g., *Epipremnum aureum*, *Monstera deliciosa*, and *Calathea makoyana*). Duan *et al.* [14] applied the octree-searching to segment 3D wheat into individual organs, and the phenotypic parameters including tiller, leaf number, height, Haun index, phyllochron, leaf length, and angle, are then extracted from the reconstructed wheat point cloud. Mccormick *et al.* [15] proposed a segmentation technique for the stem and the leaf for sorghum point clouds by meshing. Wheeler *et al.* [16] proposed a new semi-automatic approach to cluster terrestrial laser scanned data (e.g., poplar, sweet chestnut, and red oak) into meaningful sets of points for extracting plant components like internodes, petioles, and leaf-blades. Gélard *et al.* [17] segmented and classified leaves from sunflower and sorghum point clouds acquired from a multi-view imaging system with a model-based segmentation method, and they also measured the leaf areas from the segmented models. Guo *et al.* [18] employed the multi-perspective stereo vision system to generate the point clouds of *Pachira macrocarpa*, *Scindapsus*, strawberry, and tomato plant, respectively. And they segmented leaves from the point clouds via a pipeline of three steps. Koma *et al.* [19] conducted individual leaf segmentation for the terrestrial laser scanner (TLS) point cloud of a tulip tree by region growing and calculated the morphological characteristics of leaves. Mónica *et al.* [20] proposed an octree-based 3D-grid mesh method to segment leaves from a *Calathea roseopicta* point cloud scanned from a terrestrial LiDAR, and the leaves were automatically detected with a global accuracy of 93.57% in daytime and 87.34% at night, respectively. Xu *et al.* [21] designed a computer graphics-based algorithm to segment leaves from TLS point clouds of *Ehretia macrophylla*, *Crape myrtle*, and *Fatsia japonica* plants, and the precision reached 94%, 90.6%, and 88.8%, respectively. Elnashef *et al.* [22] proposed a tensor-based classification method to segment leaves from cotton, corn, and wheat point clouds scanned from a multi-view imaging system, and the average precision reached 92%, 94%, and 95%, respectively. Hu *et al.* [23] developed a 3D point cloud filtering method for leaves. After removing the outlier clusters and noise, their method was able to visualize better 3D leaf shapes and to estimate leaf areas with a higher accuracy against the classical PCA.

Some researchers extracted 2D features from plant images to facilitate leaf segmentation in 3D point clouds. Teng *et al.* [24] adopted a 2D/3D joint method to segment the target blade from simple plant point clouds. Xia *et al.* [25] combined the depth image with the 3D point cloud acquired from Kinect v1 sensor to carry out in situ leaf segmentation in greenhouse and reached a total segmentation rate of 87.97%. Chaurasia *et al.* [26] designed a clustering method based on super-pixel graph to pre-segment leaves from plant point clouds, and then used iterative closest point (ICP) matching to refine the pre-segmented results. Itakura *et al.* [27] proposed an automatic leaf segmentation method based on attribute-expanding for the plant point cloud from a multi-view imaging system with an average precision at 86.9%. They also calculated two leaf parameters—leaf area and inclination angle. Although 3D plant point clouds contain abundant phenotypic features, plant leaves are normally distributed in clusters and are sometimes seriously crowded in the canopy. Heavily overlapped leaves can cause considerable performance drop for nearly all existing segmentation methods. Therefore, it motivates us to develop a novel overlapping-free leaf segmentation approach that is not only capable of handling crowded canopies, but also can work on species with different leaf shapes.

The main contributions of this work are summarized as follows: (i) We propose a new 3D joint filtering operator to first separate occluded leaves and then segment them precisely. The operator can effectively separate leaves with different overlapping situations. (ii) By introducing the facet over-segmentation and facet-based region growing, the noise in segmentation is suppressed and each separated leaf center can grow to the complete leaf area; so that the dense canopy point cloud can be correctly segmented into a set of individual leaves. (iii) The proposed method can help to automatically calculate phenotypic traits of each single leaf (such as the area, length, and width), which shows the potential to become a highly effective tool for plant research and agricultural engineering. Experiments show that the average estimation errors of leaf area, length, and width for a point cloud of *Calathea makoyana* are merely 0.47%, 2.89%, and 4.64%, respectively. (iv) The experimental results show that the proposed method is effective in segmenting individual leaves from crowed point clouds of different plant species, and is also applicable on point clouds scanned from three kinds of 3D imaging systems. Our method obtains a point-level cover rate of 97% for *Epipremnum aureum* sample plant, 99% for *Monstera deliciosa*, 99% for *Calathea makoyana*, and 87% for *Hedera nepalensis*. At the leaf level, our method reaches an average Recall at 100.00%, Precision at 99.33%, and an average F-measure at 99.66%, respectively. Furthermore, the average speed of the segmentation costs only 12.92 seconds per plant on a desktop PC.

The paper is organized as follows. In Section 2, we describe the tools, experimental subjects, and the technical overview of the proposed method. Five preprocessing filters for removing noise and non-leaf points are presented in Section 3. The new 3D joint filtering operator is elaborated in Section 4. The complete algorithm of individual leaf segmentation based on pre-segmented leaf centers and facet over-segmentation is shown in Section 5. Section 6 shows experimental results, performance of the segmentation, as well as the parameter tuning process. In Section 7, we discuss the influence of the number of 3D joint filtering on the segmentation result, and we also apply the proposed method to help estimating leaf traits. The conclusion is drawn in Section 8.

This research differs from our previous work [4] in many aspects such as the workflow structure, the specific algorithm modules that build up the workflow, and most important of all—the ability of separating overlapping leaves in dense point clouds of plants. The only two connections with [4] are: (i) the pre-processing stage is an extended version of the counterpart in the previous work, and (ii) the facet over-segmentation technique for clustering the 3D-joint-filtered points is descended from the implementation in [4].

## 2. Materials and Methods

*2.1. Platforms and Subjects*

2.1.1. Platforms

The processing unit is a desktop PC with an Intel Core i7-7700 CPU (Intel, Santa Clara, CA, USA) and 16 GB RAM. The software environment is the VS2013 (Microsoft, Redmond, WA, USA) with the Point Cloud Library (PCL) [28], which is operated under Windows 10. In this paper, three types of imaging platforms with tripods are adopted to scan sample plants for point clouds. The first platform is a binocular stereo vision system proposed in [5] (as illustrated in Figure 1a). This stereo vision platform is consisted of two high-definition webcams (HD-3000 series, Microsoft, Redmond, WA, USA), a supporting board (LP-01, Fotomate, Jiangmen City, China) with a scale line, and a tripod (VCT-668RM, Yunteng Photographic Equipment Factory, Zhongshang City, China). The second platform is a Kinect V2 sensor [29] (Kinect V2, Microsoft, Redmond, WA, USA) that obtains depth information by capturing reflections from the projected pattern of infrared light. The Kinect sensor is mounted on the same type of tripod as used for the first stereo platform (as illustrated in Figure 1b). The third platform is a multi-view stereo vision system based on the structure from motion [30]. The platform employs a cell phone (MI 5s, MI, China) with a rear-mounted camera (IMX378, Sony, Japan) to capture multiple images for corps on an electric turntable (as illustrated in Figure 1c) from different views, respectively. Then the point cloud can be generated by importing images into a software called VisualSFM [31], which is easy to operate.

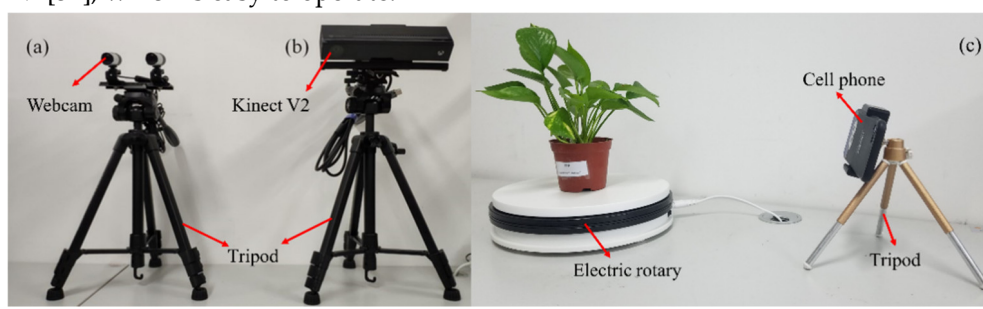

**Figure 1.** Three types of imaging platforms used in this research. (a) shows the binocular stereo vision platform containing two high-definition webcams; (b) shows the platform of the Kinect V2 sensor mounted on a tripod; (c) shows a multi-view stereo vision platform containing a cell phone, an electric turntable, and the VisualSFM software.

2.1.2. Experimental Subjects

Four types of plants are adopted as research subjects in this paper: *Epipremnum aureum*, *Monstera deliciosa*, *Calathea makoyana*, and *Hedera nepalensis*. The Kinect V2 platform is used to acquire the point cloud of an *Epipremnum aureum* sample plant. The binocular stereo vision platform is used to reconstruct the 3D point cloud of a *Monstera deliciosa* sample plant. The multi-view platform is used to obtain the point cloud of *Calathea makoyana* sample plant from 102 indoor images and the point cloud of *Hedera nepalensis* sample plant from 76 outdoor images.

*2.2. Overview of the method*

Our individual leaf segmentation approach for plant point clouds comprises of three steps. The first step is to preprocess of the original point cloud for removing noise points and non-leaf areas (e.g., stems, and background). We concatenate five different types of point cloud filters to generate pure canopy point clouds from the point clouds captured directly from imaging platforms. In the second step, we carry out 3D joint filtering for the preprocessed point cloud of plant canopy, and the filtering operator is consisted of a radius-based outlier filter and a 3D surface boundary filter. This 3D joint filtering will erode all sharp edges of a 3D surface, which resembles to the morphological erosion operation in the traditional 2D image processing. By filtering out the overlapping part among leaves in a crowded point cloud with the joint filtering operator, we are able to separate each single leaf. After this step, the point cloud of a plant canopy is divided into two parts. One part contains the remaining areas of leaves, which are mostly the center areas of original leaves. The other part is the filtered areas of leaves, the majority of which are edges of all leaves. A 3D region growing with a breadth-first searching strategy is then carried out to label each leaf center area from the remaining part with a distinctive leaf index. In the third step, facet over-segmentation is employed on the edges of leaves that were formerly filtered by the 3D joint filtering operator. Then the facets are added back to the labeled leaf centers by growing the index of leaf centers from inside to outside, and we obtain the final segmentation result for each single leaf when all facets are labeled. If the canopy is heavily crowded, we can carry out the 3D joint filtering for multiple times and then segment each leaf by growing the leaf index from the remaining center part to the outmost filtered edge part for a satisfactory result. Figure 2 uses a real segmentation example to demonstrate the overview of the proposed segmentation method in a flow diagram. In order to assure a good segmentation, the 3D joint filter is applied twice to separate individual leaves from a crowded point cloud of *Epipremnum aureum*.

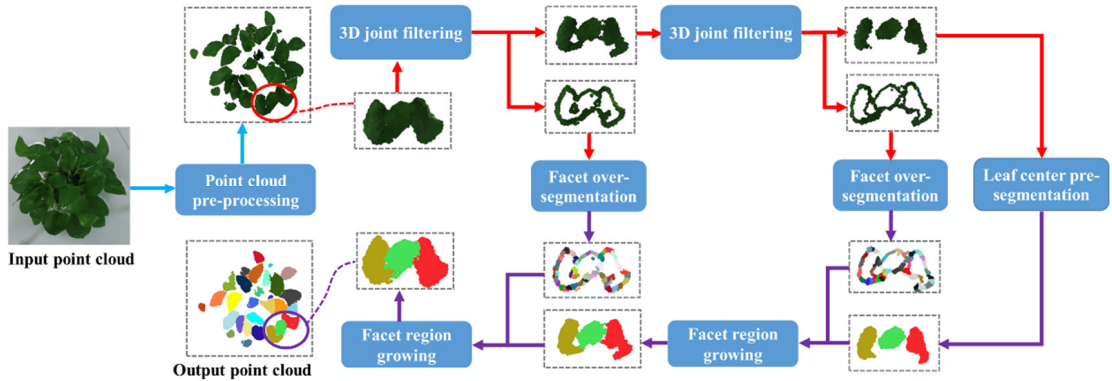

**Figure 2.** Overview of the proposed method on segmenting a point cloud of *Epipremnum aureum*. Due to the high complexity of the 3D structure of the canopy, the proposed 3D joint filtering operator are performed twice to separate heavily overlapped leaves.

## 3. Point Cloud Pre-processing

During reconstruction of a plant point cloud, an imaging system will easily import non-leaf information (e.g., pots, and ground area). And due to the limitation of imaging accuracy, the reconstructed surface of a smooth leaf in the point cloud is usually rugged. Ubiquitous imaging noise can also contaminate the point cloud with isolated points. These issues can easily mix and connect adjacent leaf surfaces that are actually separated in the real 3D space, which poses a great challenge to leaf segmentation algorithms [4]. Plants usually have various structures and blade shapes. Thus, in order to reduce non-leaf information and noise, we employed different preprocessing methods for different plant point clouds. We form a basic set containing five filtering for point cloud preprocessing. Filter I is a filter based on spatial region filtering, which removes all points outside a region defined in a 3D coordinate system from the point cloud. For example, if the plant is scanned from top and Z-axis is aligned with the direction of depth, we can remove ground points by leaving out points that do not fall into an interval of Z-axis because the canopy lies above the ground. Filter II is a radius-based outlier filter. The principle of the filter is that if the number of points in the sphere of radius $r$ centered at the query point $\mathbf{x}_q$ is lower than a threshold $n_{threshold}$, then $\mathbf{x}_q$ will be considered as an outlier and then discarded. A typical realization of Filter II is the "RadiusOutlierRemoval" function in the PCL library. The filter can remove sparse points and outliers, and is especially good at suppressing interpolated points generated by sensors and removing small plant stem areas in point clouds. Filter III is a statistical k-nearest neighbor filter. Its principle is first to calculate the average distance between the $k$-nearest neighboring points and the query point $\mathbf{x}_q$. The neighbor point whose distance to $\mathbf{x}_q$ is larger than a standard deviation will then be removed. A typical realization of this filter is the "StatisticalOutlierRemoval" function in the PCL library. Filter IV is a color-based filter. Some non-leaf information that is embedded into the canopy area in the point cloud is difficult to be removed by spatial filters; but we can distinguish non-leaf points from leaves according to the degree of greenness in RGB color because most of leaves are naturally green. Filter V is a down-sampling method. This filter creates 3D voxel grids in the point cloud, and then for each voxel, its center of gravity replaces all points in it. This filter can unify the point clouds scanned from different imaging platforms with a similar point density, which may facilitate the parameter tuning process for many point cloud based algorithms.

Figure 3 lists sets of filters used for preprocessing the point clouds of four plant types, respectively. For the point cloud of *Epipremnum aureum* sample plant, Filter I, Filter II, and Filter III are concatenated to make sure that only points belong to leaves remain in the output. Due to large leaf sizes and the sparsity in canopy structure, a concatenation of only two filters: Filter I and III, is enough to preprocess the point cloud of *Monstera deliciosa* sample plant to achieve a satisfactory output. For the dense point cloud of *Calathea makoyana* scanned from the multi-view imaging platform, Filter I, Filter IV, Filer II, and Filter V are sequentially applied. And for the point cloud of *Hedera nepalensis* sample plant, we use Filter I, Filter II, and Filter V to achieve a square cluster of leaves. If practitioners want to test our method on plant point clouds of theirs, we suggest first comparing their plants with the four types in Figure 3 on both aspects of leaf size and canopy density, and then applying the corresponding set of filters of the most similar one to theirs.

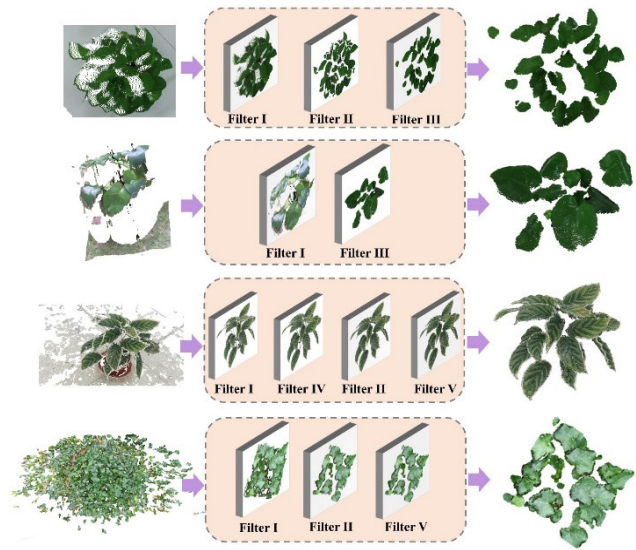

**Figure 3.** Concatenations of filters used for preprocessing the point clouds of four types of sample plants, respectively. The first row shows the filters used for processing the point cloud of *Epipremnum aureum* sample plant, including Filter I, Filter II, and Filter III. The second row demonstrates the filters used for processing the point cloud of *Monstera deliciosa* sample plant. The third row shows the filters for the point cloud of *Calathea makoyana* sample plant. The fourth row shows the concatenation of filters for processing the point cloud of *Hedera nepalensis* sample plant.

## 4. The 3D joint filtering operator

### 4.1. Defining overlapping

In general, there are two kinds of leaf overlapping in a plant point cloud. The first kind is called cross overlapping, which either stems from the contact of several curved blades from different layers in the canopy, or from the non-parallel connection among adjacent blades on a same layer. A cross overlapping example of four *Hedera nepalensis* leaves are illustrated in two views by Figure 4(a) and 4(c), respectively. The overlapping area of the local point cloud contains four highly curved leaves (Nos. 1~4). Leaf No. 3 contacts leaves No. 1 and No. 4 from different layers, and also touches the blade No. 2 at the same layer in a non-parallel way. Figure 4(b) and 4(d) show the spatial structure of the four leaves rendered in mesh from the top and the oblique view, respectively. And each red arrow is the average normal of that leaf, and all of the arrows have the same length. The second kind of overlapping is called coplanar overlapping, in which two (or more) flat and coplanar leaves connect with each other, or one leaf covers the others on a plane. A coplanar overlapping example of two

*Monstera deliciosa* leaves are illustrated in two views by Figure 4(e) and 4(g), respectively. The overlapping area contains two flat leaves. Leaf No. 5 and leaf No. 6 are on a common plane and their edges connect with each other. The shared area between the two leaves cannot be easily classified to either leaf No. 5, or No. 6 due to the ambiguity of coplanar overlapping. Figure 4(f) and 4(h) illustrate the spatial structure of the two leaves rendered in mesh from the top and the side view, respectively. And the red arrows are the respective average normals of the same length. Actually, absolute cross overlapping and coplanar overlapping are both rare in a plant point cloud. An ordinary leaf overlapping situation in the canopy we usually come across is a combination of the two overlapping types, which is highly complicated. In order to differentiate each single leaf from dense plant canopies in the real world, we propose a 3D joint filtering operator that deals with both cross overlapping and coplanar overlapping at the same time. This operator can filter out the connected areas so that we can distinguish individual leaves better.

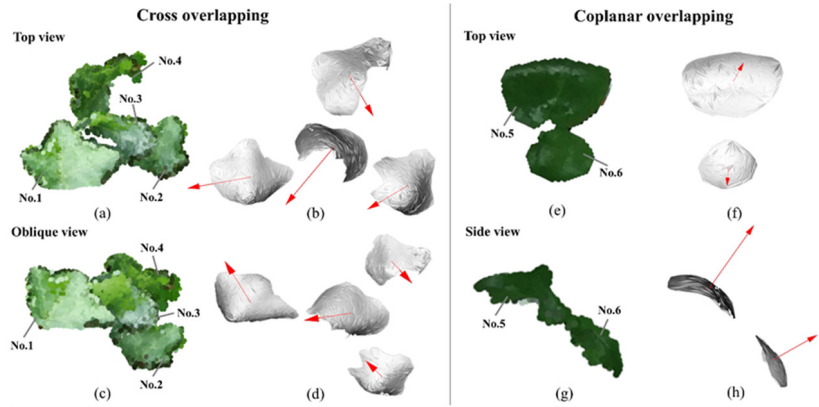

**Figure 4.** Examples of cross overlapping and coplanar overlapping in the point clouds of real plants. A cross overlapping example of four *Hedera nepalensis* leaves are illustrated in the left. (a) and (c) show the selected point cloud contains leaves Nos. 1~4 from the top and the oblique view, respectively. (b) and (d) illustrate the spatial structure rendered in mesh of (a) and (c), respectively. Each red arrow represents the unit normal of its leaf. Two leaves (Nos. 5~6) with coplanar overlapping selected from the *Monstera deliciosa* point cloud are shown in the right. (e) and (g) illustrate the overlapping point cloud from the top and the side view, respectively. (f) and (h) show the meshed spatial structure of the two leaves in mesh from two views, respectively.

### 4.2. The 3D joint filtering operator

In this sub-section, we propose a novel 3D joint filtering operator by integrating a Radius-based Outlier Filter (RBOF) and a Surface Boundary Filter (SBF).

The principle of the RBOF is similar to the Filter II used in point cloud preprocessing. Normally, the points in cross overlapping area have a lower density than the points in the leaf area. Therefore, RBOF can separate leaves by removing those sparse points in the overlapping area. It should be noted that though RBOF shares the same principle with Filter II at the preprocessing stage, the parameter configurations of the two are very different. The values of parameters $r$ and $n_{threshold}$ used for RBOF should be set according to the real density of the leaf point cloud. Figure 5 illustrates the principle of the proposed RBOF, and we can easily find that the filter is good at removing sparse outliers and cross overlapping points.

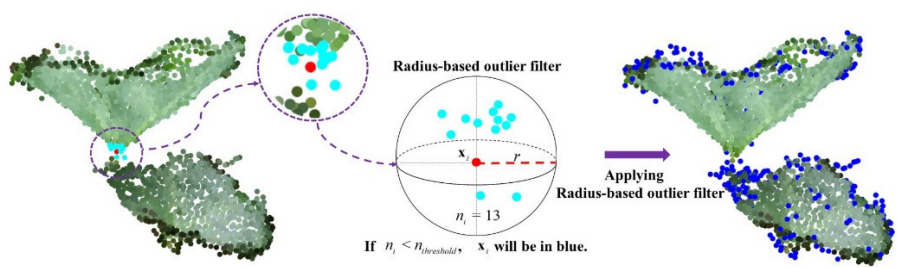

**Figure 5.** A principle demonstration of the Radius-based Outlier Filter (RBOF). The figure illustrates that how RBOF works on two *Hedera nepalensis* leaves with cross overlapping. The point $x_i$ in red is the current query point. The neighboring points in the sphere of radius $r$ centered at $x_i$ are labelled in light blue and the number of neighboring points is $n_i$. If $n_i < n_{threshold}$, the point $x_i$ is regarded as an outlier and will be labeled in dark blue later as shown in the right.

The coplanar overlapping phenomenon appears when several flat leaves are almost on the same plane in the 3D space and are connected. Hence, it comes naturally to us to design an erosion operation that functions like the morphological erosion in 2D image processing, to remove the boundaries of overlapped leaves for separation. Although the 2D erosion is a basic and easy-to-realize morphological operator in image processing, it cannot be directly extended to a surface in the 3D space. Some researchers turned to the erosion algorithm base on 3D voxels [32]. However, the voxel-based method filters all points of the surface unanimously, and alters the shape of the surface in the 3D space. If it is applied on plant point clouds, leaves will be seriously broken and the changes in shape create new problems for segmentation. The reference [33] summarized a method based on Principle Component Analysis (PCA) to extract boundary points of curved surfaces in the three-dimensional space. This approach can detect most of the surface edge points like an erosion operator in 2D. Inspired by [33], we design a Surface Boundary Filter (SBF) to extract the edge points at the overlapping area to separate leaves. The steps of SBF are as follows:

(i) Find the $k$-nearest neighborhood of the query point $x_i$ and push them into the point sets $\chi_i^k$;

(ii) Calculate the normal vector $\mathbf{n}$ of $\chi_i^k$ and the other two component vectors $\mathbf{u}$ and $\mathbf{v}$ by PCA. ($\mathbf{n} \perp \mathbf{u} \perp \mathbf{v}$);

(iii) Form a vector $\mathbf{x}_j - \mathbf{x}_i$ from the query point $\mathbf{x}_i$ and a point $\mathbf{x}_j$ in $\chi_i^k$, and the vector $(\mathbf{x}_j - \mathbf{x}_i)_{\mathbf{uv}}$ is the projection of this vector on the normal plane constructed by $\mathbf{u}$ and $\mathbf{v}$. $\theta_j$ is an angle between the vector $(\mathbf{x}_j - \mathbf{x}_i)_{\mathbf{uv}}$ and $\mathbf{u}$; it is computed by

$$\theta_j = \arccos\langle(\mathbf{x}_j - \mathbf{x}_i)_{\mathbf{uv}}, \mathbf{u}\rangle, \text{with } -\pi \leq \theta_j \leq \pi, \text{ and } j \in \{1,...,k\}. \qquad (1)$$

(iv) Sort the angle set $\Theta = \{\theta_j \mid j \in \{1,...,k\}\}$ in ascending order. If the maximum angle difference satisfies $\max(\theta_{j+1} - \theta_j) > \theta_{threshold}$, then $\mathbf{x}_i$ is viewed as a boundary point and will then be removed.

(v) After all points have performed the above steps, one iteration of filtering is completed. To achieve a satisfactory filtering result, we can repeat with several iterations on the remaining point cloud. In all experiments of this paper, we iterate three times in a single round of SBF to remove enough boundary points.

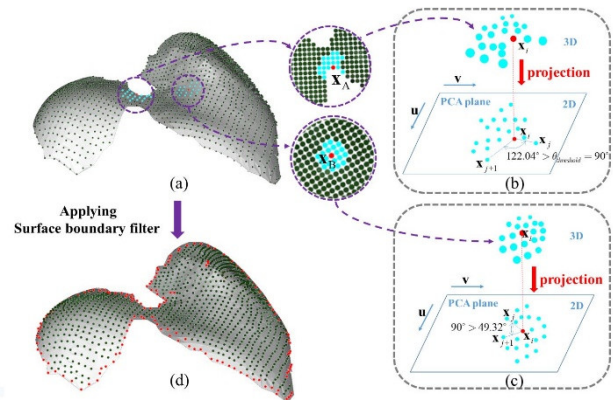

**Figure 6.** A principle demonstration of the Surface Boundary Filter (SBF) on two connected *Monstera deliciosa* leaves. In (a) we choose two points, A from the overlapping area and point B from a leaf center to show how SBF works differently on the boundary and non-boundary points. (b) illustrates that after doing PCA and projection, the maximum angle difference $\max(\theta_{j+1} - \theta_j)$ of point A is $122.04°$, which is larger than the threshold $\theta_{threshold} = 90°$. (c) shows that after doing PCA and projection, the maximum angle difference $\max(\theta_{j+1} - \theta_j)$ of point B is $49.32°$, which is smaller than the threshold. (d) shows the boundary points (painted in red) detected by just one iteration of SBF.

Figure 6 demonstrates how the SBF detects the boundary points for two *Monstera deliciosa* leaves with coplanar overlapping. In Figure 6, points A locates at the overlapping area and point B locates in a leaf center. We perform SBF respectively on the two points. The maximum angle difference of point A is larger than $\theta_{threshold}$ that is fixed at $90°$. Therefore, A is detected as a boundary point. The maximum angle difference of point B is smaller than the threshold $\theta_{threshold} = 90°$, so B is decided to be an inner point that should stay in the point cloud. Figure 6(a) shows the actual positions of the point A and point B in the meshed cloud, and the red points in Figure 6(d) are the boundary points detected by one iteration of SBF. In just one iteration, the proposed edge filter perfectly captures the boundary of the connected point cloud, and it even recognizes several outlier points that lie on the peak of the right leaf surface in Figure 6(d).

In order to handle cross overlapping, coplanar overlapping, and diverse combinations of the two situations. The RBOF and the SBF are concatenated to form the joint 3D filtering operator. Generally, the RBOF is effective in separating leaves with cross overlapping, while SBF is designed specifically for dealing with leaves with coplanar overlapping. Figure 7 shows the filtering results of RBOF and SBF, respectively. The left part of Figure 7 illustrates the results of RBOF on two cross-overlapped *Hedera nepalensis* leaves (Nos. 1, 2) from two different views. Both leaves have funnel-like 3D structure. The blue points in Figures 7(c) and 7(d) are overlapped points and outliers detected by RBOF. Figures 7(e) and 7(f) are the filtered output of RBOF from 7(a) and 7(b), respectively. And we can observe that after removing those blue points, the leaf No. 1 and leaf No. 2 are now well separated. The right part of Figure 7 shows the result of SBF with three iterations on two coplanar-overlapped *Monstera deliciosa* leaves (Nos. 3, 4) from two different views. The red points in Figures 7(i) and 7(j) are the boundary points and outliers detected by SBF with a standard three iterations. Figures 7(k) and 7(l) are the filtered output of SBF from 7(g) and 7(h), respectively. Just like the erosion operator used in 2D images, the SBF perfectly filters out the overlapping area of leaves and separates leaf No.3 from leaf No.4.

The pseudocode for 3D joint filtering is listed in Table 1, in which the RBOF runs from line 1 to line 9, and the SBF covers line 10 to line 25. The input point cloud $\chi$ can be divided into the remaining area $\chi_C$ and the filtered area $\chi_B$ after the 3D joint filtering. The point set $\chi_C$ is only consisted of those leaf centers that are well separated by the proposed 3D joint filtering operator, while $\chi_B$ will be over-segmented into small facets and then be classified into different leaves according to labels of leaf centers by region growing in the next sub-section. For plant canopies with serious leaf overlapping, we can run the algorithm in Table 1 for multiple times to obtain satisfactory leaf center areas. The final individual leaf segmentation result is obtained by growing the labels of $\chi_C$ to all facets of $\chi_B$ from the very inside to the outside.

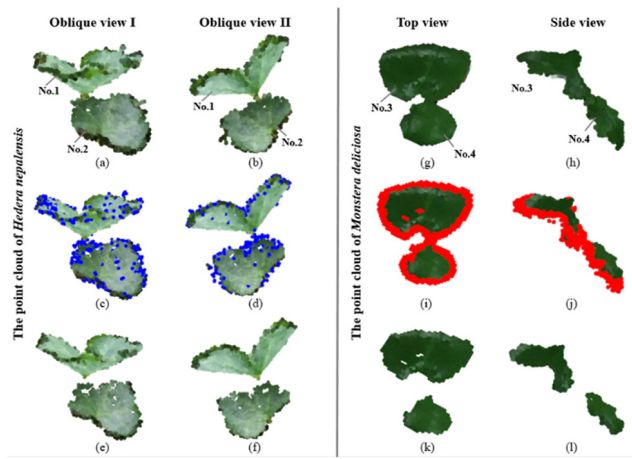

**Figure 7.** The filtering results of RBOF and SBF on two kinds of leaf overlapping, respectively. The left part illustrates the results of RBOF on two cross-overlapped *Hedera nepalensis* leaves (Nos. 1, 2), from two different views. The right part shows the results of SBF with three iterations on two coplanar-overlapped *Monstera deliciosa* leaves (Nos. 3, 4) from two different views. (a) and (b) show the original point cloud of the two cross-overlapped leaves from two views, respectively. The outliers and sparse points are detected as blue points in (c) and (d) by RBOF. (e) and (f) are the point cloud after RBOF for (a) and (b), respectively. (g) and (h) show the original point cloud of the two cross-overlapped leaves from two views, respectively. The boundary points and spurious points are detected as red points in (i) and (j) by SBF. (k) and (l) are the point cloud after SBF for (a) and (b), respectively

**Table 1.** Pseudocode for 3D joint filtering.

**Algorithm 1**: 3D joint filtering

**Inputs:** Point cloud $\chi$

**Parameters:** $n_{threshold}$, $r$, $k$, $\theta_{threshold} = 90°$ and $n_{iter} = 3$

**Outputs:** Point cloud after filtering $\chi_C$; the filtered point cloud part $\chi_B$

RBOF:
1. $\chi_C \leftarrow \varnothing$, and $\chi_B \leftarrow \varnothing$
2. **for** each point $\mathbf{x}_i$ in $\chi$ **do**
3.     Count the number of points in the sphere of radius $r$ centered at $\mathbf{x}_i$ as $n$.
4.     **if** $n < n_{threshold}$ **then**
5.         $\chi_B.push\_back(\mathbf{x}_i)$.
6.     **else**
7.         $\chi_C.push\_back(\mathbf{x}_i)$.
8.     **end if**
9. **end for**

SBF:
10. **while** $n_{iter} > 0$ **do**
11.     **for** each point $\mathbf{x}_i$ in $\chi_C$ **do**
12.         $\Theta \leftarrow \varnothing$.
13.         Initialize $\mathbf{x}_i$'s $k$-nearest neighbors $\chi_i^k$.
14.         Compute the three principal components of $\chi_i^k$ as $\mathbf{u}_i$, $\mathbf{v}_i$, and $\mathbf{n}_i$, whose corresponding eigenvalues satisfy $\lambda_\mathbf{u} \geq \lambda_\mathbf{v} \geq \lambda_\mathbf{n}$.
15.         **for** each point $\mathbf{x}_j$ in $\chi_i^k$ **do**
16.             Use equation (1) to compute $\theta_j$.
17.             $\Theta.push\_back(\theta_j)$
18.         **end for**
19.         Sort all $\theta s$ in $\Theta$ in ascending order.
20.         **if** $max(\theta_{j+1} - \theta_j) > \theta_{threshold}$ **do**
21.             $\chi_B.push\_back(\mathbf{x}_i)$, and $\chi_C.erase(\mathbf{x}_i)$.
22.         **end if**
23.     **end for**
24.     $n_{iter} \leftarrow n_{iter} - 1$.
25. **end while**
26. Set $\chi \leftarrow \chi_C$ if 3D joint filtering is to be applied multiple times.

## 5. Segmentation

### 5.1. Leaf center pre-segmentation

After applying the 3D joint filtering operator for once or multiple times, most of the remaining points of canopy are located in the central areas of the leaves; and no overlapping should exist among nearby leaf centers. Therefore, now we can label and pre-segment the left point cloud $\chi_C$ by 3D region growing. The main idea of 3D region growing is as follows. Firstly, we randomly choose an unlabeled point from $\chi_c$, and mark the point with a new leaf label. Then start from this point and search for unlabeled point within a small neighborhood. If unlabeled points are found, we mark these points with the same label as the starting point and push them into a queue from the back. Secondly, we pop out a point from the front of the queue, and the point will be used as a new search point. The first step will be repeated until the queue is empty; after that, we think a leaf center is completely pre-segmented. Lastly, the former two steps will be repeated until all points in the point cloud $\chi_c$ are labeled. Essentially, the pre-segmentation of each leaf center is a kind of breadth-first 3D region growing. The algorithm is summarized in Table 2, and the output is a set of labeled leaf points, defined by $\mathcal{F}_c$.

**Table 2.** Pseudocode for leaf center pre-segmentation.

| | |
|---|---|
| **Algorithm 2**: leaf center pre-segmentation by 3D region growing. | |
| **Input:** The point cloud that contains central areas of leaves: $\chi_c$ | |
| **Parameter:** $d_1$ | |
| **Output:** The collection of individual leaves $\mathcal{F}_c$ (only central areas). | |
| 1 | Set label $L=0$. |
| 2 | **for** each unlabeled point $\mathbf{x}_i$ in $\chi_c$ **do** |
| 3 |    Establish a queue $\mathcal{A} \leftarrow \varnothing$ for Breadth-First Searching, and set $\mathcal{A}.push\_back(\mathbf{x}_i)$. |
| 4 |    Set $\mathbf{x}_i$ as the starting point of a new individual leaf, and give $\mathbf{x}_i$ a label $L$. |
| 5 |    $\mathcal{A}.push\_back(\mathbf{x}_i)$. |
| 6 |    **while** $\mathcal{A} \neq \varnothing$ **do** |
| 7 |      $\mathbf{x}_j = \mathcal{A}.pop\_front$. |
| 8 |      **for** each point $\mathbf{x}_j$ **do** |
| 9 |        **Repeat** |
| 10 |          **if** a nearby point $\mathbf{x}_k$ is unlabeled and $\|\mathbf{x}_j - \mathbf{x}_k\| \leq d_1$ **then** |
| 11 |            $\mathcal{A}.push\_back(\mathbf{x}_k)$. |
| 12 |            Label $\mathbf{x}_k$ with $L$. |
| 13 |          **end if** |
| 14 |        **Until** all nearby points of $\mathbf{x}_j$ are visited. |
| 15 |      **end for** |
| 16 |    **end while** |
| 17 |    $L \leftarrow L+1$. |
| 18 | **end for** |
| 19 | Collect all labels of $\chi_c$ as pre-segmented individual leaves $\mathcal{F}_c$. |

*5.2. Facet over-segmentation for filtered point clouds*

The whole point cloud can be divided into the leaf center areas $\chi_c$ and the boundary area $\chi_B$ that contains overlapping parts and outliers after 3D joint filtering. Although algorithm 2 in sub-section V-A has labeled and separated the central areas of different leaves, respectively, $\chi_B$ is yet to be segmented. It is extremely difficult to assign the points of $\chi_B$ with labels of nearby leaf centers because these edge points from the overlapping areas have obvious ambiguity. Therefore, it poses a great challenge to leaf segmentation methods, and sometimes even a human is uncapable of carrying out fully correct segmentations. Similar to the segmentation method based on super-pixels in 2D image processing [34], the facet over-segmentation algorithms for point clouds [4], [13] are able to segment a point cloud into flat clusters in each of which points have similar spatial characteristics. Since the facet over-segmentation works on a larger scale, it not only reduces the number of features but also avoids direct segmentation based on individual point features that are vulnerable to noise. Moreover, if several nearby over-segmented facets have similar overall characteristics, they can by further combined into a larger plane. Therefore, facet over-segmentation seems to be an excellent tool to address the high ambiguity in separation of overlapped point clouds.

We employed the algorithm used in [4] to over-segment the filtered area $\chi_B$ and the over-segmentation method include three steps. First, we calculate the spatial characteristics of all points in $\chi_B$, such as smoothness and normal. Second, by deploying some seed points from the point cloud based on calculated spatial characteristics we can coarsely segment the point cloud into facets. At last, we refine the coarse facets by the local K-means clustering. The method was purely unsupervised, and the generation for coarse facets helps to select the value K in the last step. In the next sub-section, labels of leaf centers will grow to segmented facets from inside to outside for the final individual leaf segmentation.

*5.3. Leaf segmentation by facet region growing*

The central areas of leaves are segmented and labeled through point-based 3D region growing, and the boundary areas are consisted of many facets after over-segmentation. Now we grow labels of leaf centers to the outside facets. During the growth process, the facet region growing contains several steps. The first step is to traverse all points on a labeled leaf center area; if a point in the leaf center is adjacent to an unlabeled facet, all points belong to this facet will be assigned the same label (and also painted with the same color in segmentation results) with that leaf center. This step continues until all leaf centers do not expand any more. If unlabeled facets in $\chi_B$ remain, they are not adjacent to any leaf centers. This also means the unlabeled facets should represent new leaf areas, and they should be assigned new leaf labels. Finally, breadth-first facet region growing is performed on each new leaf area, in which every new leaf area will grow its label to its adjacent facets that are still unlabeled; this last step continues until all facets in $\chi_B$ are labeled. The pseudocode for final leaf segmentation by facet region growing is shown in Table 3.

**Table 3.** Pseudocode for final leaf segmentation by facet region growing.

| | |
|---|---|
| **Algorithm 3:** Final leaf segmentation by facet region growing | |
| **Input:** Filtered part $\chi_B$ from the latest 3D joint filtering process; The collection of leaf set $\mathcal{F}_C$ (only central areas). | |
| **Parameters:** Please refer to [4] for parameters for facet over-segmentation. | |
| **Output:** The collection of individual leaves $\mathcal{F}_C$. | |
| 1 | Utilize Algorithm 2 in [4] to generate facet set $\mathcal{F}_B$ by over-segmenting $\chi_B$. |
| 2 | **Repeat** |
| 3 |   **for** each leaf area with a label $L_i$ in $\mathcal{F}_C$ **do** |
| 4 |     **for** each facet $f_j$ in $\mathcal{F}_B$ **do** |
| 5 |       **if** $f_j$ is adjacent to $L_i$ **then** |
| 6 |         Extend $\mathcal{F}_C$ by growing $L_i$ to $f_j$. |
| 7 |         Delete $f_j$ from $\mathcal{F}_B$. |
| 8 |       **end if** |
| 9 |     **end for** |
| 10 |   **end for** |
| 11 | **until** $\mathcal{F}_C$ does not grow anymore |
| 12 | **while** $\mathcal{F}_B \neq \varnothing$ **do** |
| 13 |   Set $f_j$ as a new leaf in $\mathcal{F}_C$, and assign it a new label $L_{new} = \max(L_i) + 1$. |
| 14 |   Delete $f_j$ from $\mathcal{F}_B$. |
| 15 |   Grow $L_{new}$ to adjacent facets $f_m s$ in $\mathcal{F}_B$ with Breadth-First Searching. |
| 16 |   Delete $f_m s$ from $\mathcal{F}_B$. |
| 17 | **end while** |

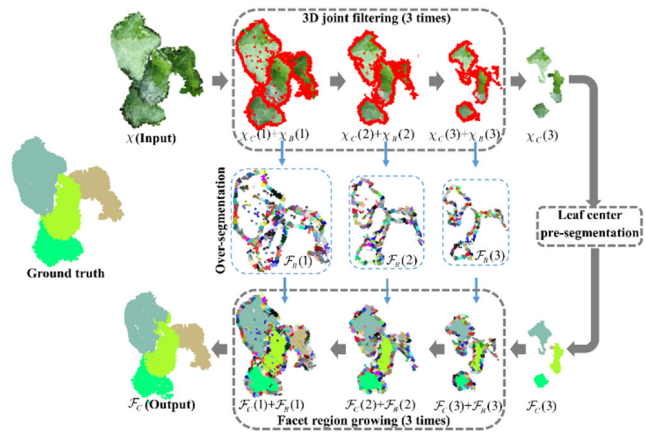

**Figure 8.** A complete individual leaf segmentation demonstration on a point cloud that contains four overlapping *Hedera nepalensis* leaves. Due to its serious overlapping effect, the proposed 3D joint filtering operator are performed for three times in a row. Facet region growing are also carried out for three times because there are three filtered point cloud parts.

Figure 8 illustrates the complete individual leaf segmentation process on a crowded point cloud that contains four overlapping *Hedera nepalensis* leaves. In the first row, the test point cloud is filtered three times by the 3D joint filtering operator to obtain a completely separated leaf center sets $\chi_C(3)$ for four leaves. Then, after the pre-segmentation with point-based region growing, the central areas of the leaves are assigned their own labels, respectively. The pre-segmentation result is a leaf point set $\mathcal{F}_C(3)$, which then expands toward the over-segmented facets from inside to outside, with the order of $\mathcal{F}_B(3)$, $\mathcal{F}_B(2)$, and $\mathcal{F}_B(1)$. The proposed method achieves a satisfactory result on the *Hedera nepalensis* point cloud, which is very close to the manually labeled ground truth shown in the leftmost part of Figure 8. During 3D joint filtering, some inner points in the leaf are removed as outliers, and some small leaf areas are even filtered as boundary; however, these points will not affect the facet over-segmentation and the facet region growing that follows. The reason is that the growth of labels carried out from center to the outside areas is performed in the three-dimensional space, making the isolated facets in the real leaf surface to be easily engulfed in growing and to output a correct segmentation. Small leaves can be easily removed in multiple rounds of 3D joint filtering, however, they will be discovered and labeled in the facet region growing stage that adds back the filtered point set to existing leaf centers.

## 6. Experiments

In this section, we first demonstrate some qualitative results for the proposed method on four types of plant point clouds. Then, we evaluate the performance of results on both point-level and leaf-level quantitative measures. A comparison among the proposed method, our previous method [4], and a state-of-the-art segmentation method [19] is also provided to prove its effectiveness.

*6.1. Qualitative segmentation results*

The qualitative segmentation results of our method for four types of plants are shown in Figure 9. We demonstrate each segmentation result from three different views, and our results are very close to the ground truth. It should be noted that, for the *Hedera nepalensis* point cloud with serious leaf overlapping, our method still achieves a satisfactory result.

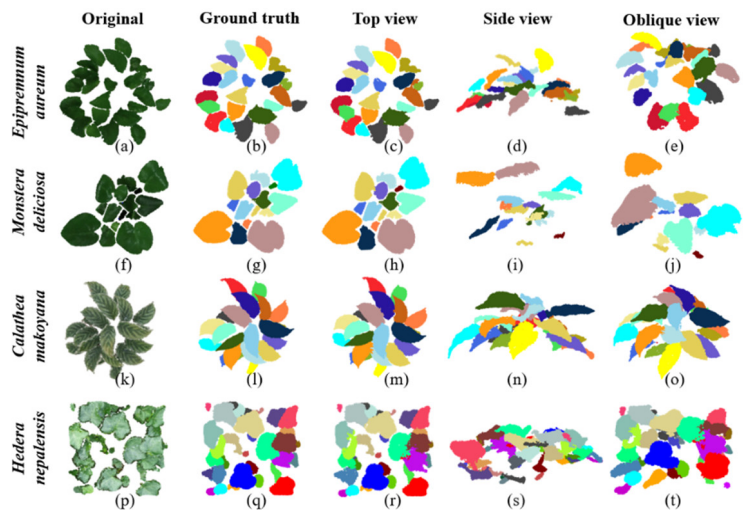

**Figure 9.** Qualitative demonstration of segmentation results of the point clouds of four types of plants, respectively. The first row is the result of the *Epipremnum aureum* sample plant. The second row is the result of the *Monstera deliciosa* sample plant. The third row is the result of the *Calathea makoyana* sample plant. The last row is the result of the *Hedera nepalensis* sample plant. The first column shows the pre-processed canopy point clouds of four plants. The second column illustrates the ground truth by manual segmentation. The third column shows the segmentation result for each point cloud with the proposed method from the top view, respectively. The fourth column demonstrates the segmentation results with the proposed method from the side view. The fifth column shows the segmentation results with the proposed method from an oblique view.

*6.2. Quantitative measures for performance evaluation*

Figure 10 uses a leaf from the *Hedera nepalensis* sample point cloud as example to explain the quantitative measures defined for evaluating algorithm performances. Figure 10(a) is the ground truth of a piece of leaf painted in original colors. The point cloud shown in Figure 10(b) is the algorithm's segmentation result for this leaf. By contrasting Figure 10(a) with 10(b), Figure 10(c) shows the contrasted result painted in different colors. If a point in the ground truth is correctly labeled in segmentation, then we call it a correct point and label it with blue color in Figure 10(c). The red area in Figure 10(c) stands for the ground truth area that is missing in an algorithm segmentation; therefore, the points in such areas are called missing points. The yellow region refers to the segmented points that do not belong to ground truth, and such points are called the false points.

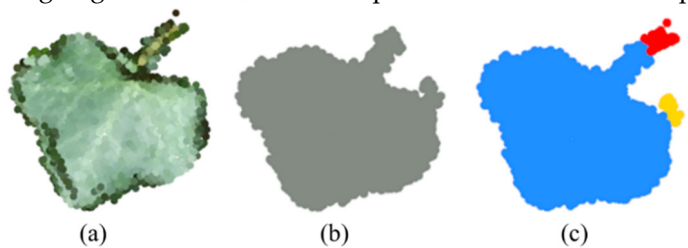

**Figure 10.** An example for explaining the quantitative measures for evaluating the segmentation result at the point level. (a) is the ground truth of a piece of *Hedera nepalensis* leaf painted with real colors. The gray point cloud in (b) is the segmentation result of the same leaf by the algorithm. In (c), the blue points are the correct points; the red points are the missing points, and yellow points are the false points.

To better analyze the results of leaf segmentation algorithms in the quantitative aspect, we devise a new index for assessing the precision of leaf segmentation. Cover_rate, defined as the ratio of the number of points $Num\_blue$ in the common area of both segmentation result and ground truth to the number of all ground truth points $Num\_gt$. The equation of cover rate is as follows:

$$Cover\_rate = \frac{Num\_blue}{Num\_gt} \times 100\%. \tag{2}$$

Meanwhile, we also introduce several quantitative measures defined in [4] to comprehensively evaluate the algorithm performance at leaf level. True Positive (TP): if a segmented leaf region covers more than 70% of the total number of the points of that real leaf, the segmented leaf is then regarded as a TP. False Positive (FP): if two leaves are falsely segmented by the same segmentation region, we regard it to be an FP. And if three leaves are falsely connected into one segmentation, then there are two FPs. False Negative (FN): If more than 70% points of a real leaf is not covered by any segmentation, then we regard it as an FN. Because we only process pure canopy point clouds, non-leaf points do not exist in ground truth, which causes the absence of the True Negative (TN). Based on the above definitions of TP, FP, and FN, we calculate three metrics on the leaf level including Recall, Precision, and F-measure to quantitatively evaluate the proposed algorithm on the leaf-level. The three metrics are defined as follows:

$$\text{Recall} = \frac{\text{TPs}}{\text{TPs} + \text{FNs}} \times 100\%. \tag{3}$$

$$\text{Precision} = \frac{\text{TPs}}{\text{TPs} + \text{FPs}} \times 100\%. \tag{4}$$

$$\text{F-measure} = \frac{2\text{TPs}}{2\text{TPs} + \text{FPs} + \text{FNs}} \times 100\%. \tag{5}$$

*6.3. Quantitative experimental results and comparison*

We perform quantitative evaluations for segmentation results at both point level and leaf level. Figures 11-14 demonstrate color labeled point-level quantitative measures, including correct points (Cp), result points (Rp), missing points (Mp), and false points (Fp), for each single leaf in the four sample point clouds. In Figures 11 to 14, the columns that with label "(a)" contain the ground truth of individual leaves rendered by the real colors in their original point clouds; the columns that with label "(b)" show the results from the proposed leaf segmentation algorithm; the columns that with label "(c)" demonstrate color labeled point-level measures in the same way as Figure 10(c). Table 4 lists all detailed values of point-level measures on each segmented leaf of *Epipremnum aureum*, *Monstera deliciosa*, *Calathea makoyana*, and *Hedera nepalensis* sample plants, respectively. In Table 4, result points (Rp) represent the number of points contained in the segmented area for a single leaf, and other measures are defined as the same in sub-section VI-B. On the four different types of plant point clouds with leaf overlapping, our method reaches full coverage for a majority of individual leaves; and in most of cases, the blue areas formed by correct points are identical to their respective ground truth areas. Therefore, our method has high effectiveness and wide applicability.

Figure 15 illustrates the quantitative and qualitative comparisons across methods [4], [19], and the proposed method on the four different plant point clouds. In Figure 15, the first and second rows show the pre-processed point clouds, and the segmentation ground truth from the top view, respectively. The third and the fourth rows illustrate the segmentation results of [19], and [4], respectively. The last row shows the segmentation results with the proposed method. The number of TPs and the average cover rate are also provided for each segmented point cloud on the upper-right corner and the lower-right corner, respectively. It is evident that the proposed method is superior than the methods [4] and [19] both on the number of successfully segmented leaves and the average segmentation accuracy.

Table 5 compares the leaf-level quantitative measures of the segmentation results of [4], [19], and the proposed method for the four different plants, respectively. The average Precision, average Recall, and average F-measure of our method for all four plant point clouds are all highest among the compared, reaching 99.33%, 100%, and 99.66%, respectively. We also record the processing time of the proposed method on our software platform. The fast point cloud costs 1.963s, and the slowest costs 18.902s. The average processing time is about 13 seconds, satisfying quasi real-time requirement. Although the method [19] is the fastest, it obtains the worst segmentation result in most cases.

Table 4. Detailed values of the point-level measures on each segmented leaf of the four sample plants.

| Leaf index | Num_gt | Rp | Mp | Fp | Num_blue | Cover_rate | Leaf index | Num_gt | Rp | Mp | Fp | Num_blue | Cover_rate |
|---|---|---|---|---|---|---|---|---|---|---|---|---|---|
| *Epipremnum aureum* | | | | | | | 10 | 1569 | 1569 | 0 | 0 | 1569 | 100% |
| 1 | 771 | 771 | 0 | 0 | 771 | 100% | 11 | 1059 | 1059 | 0 | 0 | 1059 | 100% |
| 2 | 449 | 449 | 0 | 0 | 449 | 100% | 12 | 1322 | 1322 | 0 | 0 | 1322 | 100% |
| 3 | 494 | 494 | 0 | 0 | 494 | 100% | 13 | 1278 | 1278 | 0 | 0 | 1278 | 100% |
| 4 | 673 | 673 | 0 | 0 | 673 | 100% | 14 | 759 | 759 | 0 | 0 | 759 | 100% |
| 5 | 280 | 280 | 0 | 0 | 280 | 100% | 15 | 1370 | 1370 | 0 | 0 | 1370 | 100% |
| 6 | 382 | 382 | 0 | 0 | 382 | 100% | 16 | 1532 | 1569 | 2 | 39 | 1530 | 99.9% |
| 7 | 337 | 337 | 0 | 0 | 337 | 100% | 17 | 1299 | 1314 | 0 | 15 | 1299 | 100% |
| 8 | 681 | 749 | 0 | 68 | 681 | 100% | 18 | 654 | 654 | 0 | 0 | 654 | 100% |
| 9 | 791 | 782 | 9 | 0 | 782 | 98.9% | 19 | 1504 | 1415 | 92 | 3 | 1412 | 93.9% |
| 10 | 494 | 494 | 0 | 0 | 494 | 100% | 20 | 1641 | 1568 | 73 | 0 | 1568 | 95.6% |
| 11 | 831 | 856 | 0 | 25 | 831 | 100% | 21 | 1428 | 1428 | 0 | 0 | 1428 | 100% |
| 12 | 632 | 632 | 0 | 0 | 632 | 100% | *Hedera nepalensis* | | | | | | |
| 13 | 361 | 353 | 8 | 0 | 353 | 97.8% | 1 | 1636 | 1636 | 0 | 0 | 1636 | 100% |
| 14 | 571 | 571 | 0 | 0 | 571 | 100% | 2 | 380 | 380 | 0 | 0 | 380 | 100% |
| 15 | 442 | 442 | 0 | 0 | 442 | 100% | 3 | 1934 | 1919 | 38 | 23 | 1896 | 98.0% |
| 16 | 623 | 607 | 18 | 2 | 605 | 97.1% | 4 | 828 | 828 | 0 | 0 | 828 | 100% |
| 17 | 521 | 521 | 0 | 0 | 521 | 100% | 5 | 1626 | 1953 | 107 | 434 | 1519 | 93.4% |
| 18 | 526 | 614 | 0 | 83 | 526 | 100% | 6 | 566 | 617 | 0 | 51 | 566 | 100% |
| 19 | 524 | 524 | 0 | 0 | 524 | 100% | 7 | 528 | 373 | 160 | 5 | 368 | 69.7% |
| 20 | 489 | 460 | 78 | 49 | 411 | 84.1% | 8 | 1992 | 1904 | 146 | 58 | 1846 | 92.7% |
| 21 | 222 | 155 | 67 | 0 | 155 | 69.8% | 9 | 2223 | 2312 | 0 | 89 | 2223 | 100% |
| 22 | 320 | 320 | 0 | 0 | 320 | 100% | 10 | 99 | 99 | 0 | 0 | 99 | 100% |
| 23 | 482 | 482 | 0 | 0 | 482 | 100% | 11 | 461 | 239 | 222 | 0 | 239 | 51.8% |
| *Monstera deliciosa* | | | | | | | 380 | 270 | 110 | 0 | 270 | 71.1% | 12 |
| 1 | 873 | 872 | 2 | 1 | 871 | 99.8% | 13 | 1367 | 1395 | 82 | 110 | 1285 | 94.0% |
| 2 | 2075 | 2075 | 0 | 0 | 2075 | 100% | 14 | 871 | 890 | 0 | 19 | 871 | 100% |
| 3 | 363 | 364 | 0 | 1 | 363 | 100% | 15 | 1741 | 1538 | 264 | 61 | 1477 | 84.8% |
| 4 | 178 | 178 | 0 | 0 | 178 | 100% | 16 | 1538 | 1383 | 155 | 0 | 1383 | 89.9% |
| 5 | 156 | 156 | 0 | 0 | 156 | 100% | 17 | 1206 | 1145 | 61 | 0 | 1145 | 94.9% |
| 6 | 754 | 754 | 0 | 0 | 754 | 100% | 18 | 231 | 231 | 0 | 0 | 231 | 100% |
| 7 | 1925 | 1925 | 0 | 0 | 1925 | 100% | 19 | 195 | 139 | 56 | 0 | 139 | 71.3% |
| 8 | 590 | 590 | 0 | 0 | 590 | 100% | 20 | 981 | 961 | 49 | 29 | 932 | 95.0% |
| 9 | 350 | 343 | 8 | 1 | 342 | 97.7% | 21 | 1575 | 1579 | 46 | 50 | 1529 | 97.1% |
| 10 | 62 | 62 | 0 | 0 | 62 | 100% | 22 | 154 | 154 | 0 | 0 | 154 | 100% |
| 11 | 370 | 370 | 0 | 0 | 370 | 100% | 23 | 117 | 109 | 8 | 0 | 109 | 93.2% |
| 12 | 1309 | 1309 | 0 | 0 | 1309 | 100% | 24 | 785 | 945 | 59 | 219 | 726 | 92.5% |
| 13 | 459 | 459 | 0 | 0 | 459 | 100% | 25 | 1793 | 1825 | 0 | 32 | 1793 | 100% |
| 14 | 103 | 103 | 0 | 0 | 103 | 100% | 26 | 456 | 433 | 23 | 0 | 433 | 95.0% |
| 15 | 91 | 91 | 0 | 0 | 91 | 100% | 27 | 1137 | 1173 | 13 | 49 | 1124 | 98.9% |
| *Calathea makoyana* | | | | | | | 233 | 233 | 0 | 0 | 233 | 100% | 28 |
| 1 | 1483 | 1483 | 0 | 0 | 1483 | 100% | 29 | 850 | 818 | 42 | 10 | 808 | 95.1% |
| 2 | 1485 | 1485 | 0 | 0 | 1485 | 100% | 30 | 106 | 106 | 0 | 0 | 106 | 100% |
| 3 | 1489 | 1556 | 4 | 71 | 1485 | 99.7% | 31 | 386 | 310 | 76 | 0 | 310 | 80.3% |
| 4 | 925 | 927 | 0 | 2 | 925 | 100% | 32 | 381 | 351 | 30 | 0 | 351 | 92.1% |
| 5 | 1713 | 1713 | 0 | 0 | 1713 | 100% | 33 | 822 | 923 | 0 | 101 | 822 | 100% |
| 6 | 1233 | 1233 | 0 | 0 | 1233 | 100% | 34 | 103 | 103 | 0 | 0 | 103 | 100% |
| 7 | 1706 | 1706 | 0 | 0 | 1706 | 100% | 35 | 240 | 237 | 3 | 0 | 237 | 98.8% |
| 8 | 1606 | 1617 | 0 | 11 | 1606 | 100% | 36 | 167 | 140 | 27 | 0 | 140 | 83.8% |
| 9 | 1365 | 1365 | 0 | 0 | 1365 | 100% | 37 | 243 | 0 | 243 | 0 | 0 | 0% |

Table 5. Comparison of the leaf-level quantitative measures of the segmentation results of [4], [19], and the proposed method for the four different plants, respectively.

| Plant type | Method | TP | Reference | FP | FN | Recall | Precision | F-Measure | Average cover rate | Time cost |
|---|---|---|---|---|---|---|---|---|---|---|
| *Epipremnum aureum* | [19] | 17 | 23 | 6 | 0 | 100% | 73.9% | 85% | 73.52% | 0.41s |
|  | [4] | 21 | 23 | 2 | 0 | 100% | 91.3% | 95.45% | 89% | 13.04s |
|  | Ours | 23 | 23 | 0 | 0 | 100% | 100% | 100% | 97.72% | 18.902s |
| *Monstera deliciosa* | [19] | 14 | 15 | 1 | 0 | 100% | 93.75% | 96.77% | 93.30% | 0.27s |
|  | [4] | 14 | 15 | 1 | 0 | 100% | 93.75% | 96.77% | 93.33% | 4.85s |
|  | Ours | 15 | 15 | 0 | 0 | 100% | 100% | 100% | 99.79% | 1.963s |
| *Calathea makoyana* | [19] | 18 | 21 | 3 | 0 | 100% | 85.71% | 92.31% | 80.73% | 1.239s |
| | [4] | 18 | 21 | 3 | 0 | 100% | 85.71% | 92.31% | 85% | 45.22s |
| | Ours | 21 | 21 | 0 | 0 | 100% | 100% | 100% | 99.48% | 17.199s |
| *Hedera nepalensis* | [19] | 23 | 37 | 14 | 0 | 100% | 62.16% | 76.67% | 62.44% | 1.538s |
| | [4] | 14 | 37 | 23 | 0 | 100% | 37.84% | 54.9% | 25% | 12.833s |
| | Ours | 36 | 37 | 1 | 0 | 100% | 97.30% | 98.63% | 87.39% | 13.598s |

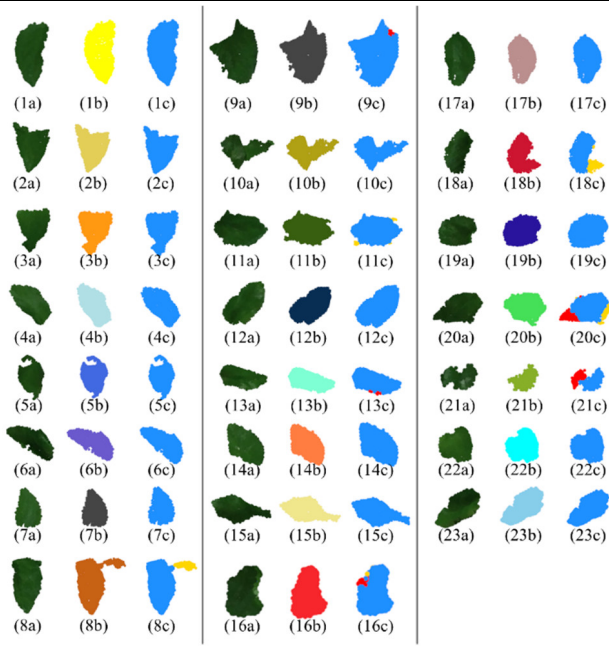

**Figure 11.** The color labeled point-level quantitative measures for all leaves in the *Epipremnum aureum* point cloud. The columns that with label "(a)" contain the ground truth of individual leaves rendered by real colors. The columns that with label "(b)" show results of the proposed segmentation algorithm, and each segmented leaf is painted with a different color; the columns that with label "(c)" demonstrate color labeled point-level measures in the same way as Figure 10(c).

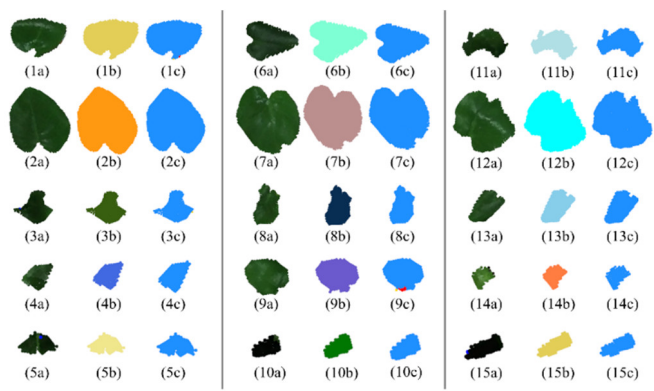

**Figure 12.** The color labeled point-level quantitative measures for all leaves in the *Monstera deliciosa* point cloud. The columns that with label "(a)" contain the ground truth of individual leaves rendered by real colors. The columns that with label "(b)" show results of the proposed segmentation algorithm, and each segmented leaf is painted with a different color; the columns that with label "(c)" demonstrate color labeled point-level measures in the same way as Figure 10(c).

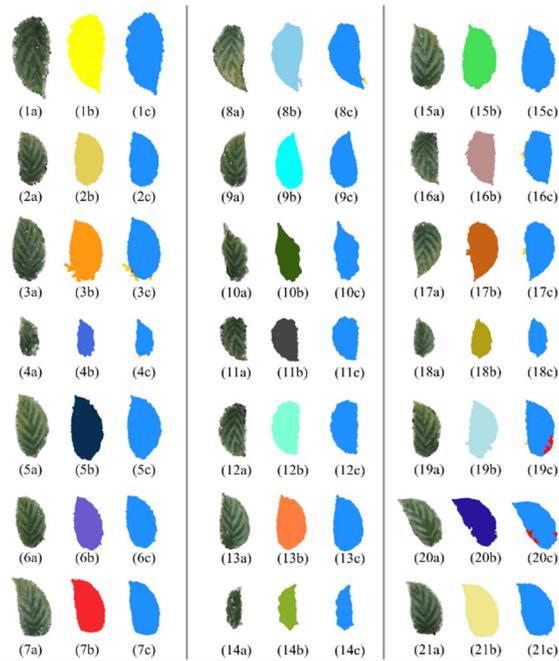

**Figure 13.** The color labeled point-level quantitative measures for all leaves in the *Calathea makoyana* point cloud. The columns that with label "(a)" contain the ground truth of individual leaves rendered by real colors. The columns that with label "(b)" show results of the proposed segmentation algorithm, and each segmented leaf is painted with a different color; the columns that with label "(c)" demonstrate color labeled point-level measures in the same way as Figure 10(c).

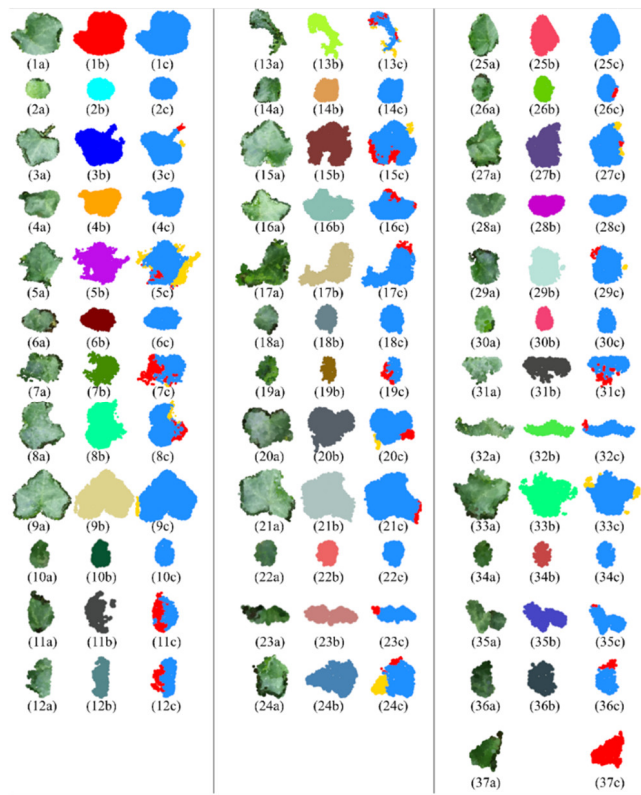

**Figure 14.** The color labeled point-level quantitative measures for all leaves in the *Hedera nepalensis* point cloud. The columns that with label "(a)" contain the ground truth of individual leaves rendered by real colors. The columns that with label "(b)" show results of the proposed segmentation algorithm, and each segmented leaf is painted with a different color; the columns that with label "(c)"

demonstrate color labeled point-level measures in the same way as Figure 10(c). Please be noted that (37b) is missing because the method false classify the leaf (24a) and leaf (37a) as one segment shown in (24b).

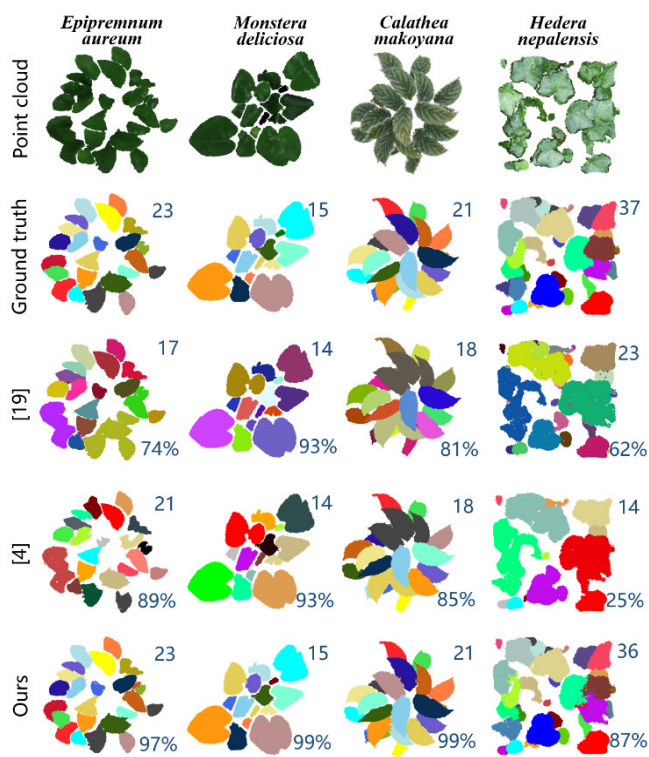

**Figure 15.** The comparison of segmentation results across [4], [19], and the proposed method on the four different plant point clouds. The first row shows the pre-processed point clouds from the top view. The second row illustrates the ground truth from the top view, and the number on the upper-right corner of each ground truth is the real number of leaves in the point cloud. The third row illustrates the segmentation results of [19], and the fourth row shows the results of [4]. The last row shows the segmentation results by the proposed method. The number of TPs and the average cover rate are also provided for each segmented point cloud on the upper-right corner and the lower-right corner, respectively.

*6.4. Parameter tuning*

The proposed segmentation method for individual leaves relies on local geometric features of Euclidean distances among points to carry out leaf labeling. Although our method is able to effectively segment leaf point clouds scanned from several kinds of sensors and imaging systems, we need to adjust the parameters separately for each of them. To facilitate researchers and practitioners on revisit of our method, we try to associate the parameters with the inherent coefficients of the point clouds such as the average spacing and the average number of points in a spherical region. Table 6 lists the average spacing values and the average number of neighborhood points in spheres with different radii for the four point clouds, respectively. Table 7 lists the configurations of the parameters of our method employed on the four types of plants; most of parameters are fixed by referring to the coefficient data listed in Table 6. According to our experience, when the value of $r$ is about four to five times to the average spacing, a good filtering result will be obtained. The value of $n_{threshold}$ should be related to the density of the canopy and the degree of overlapping among leaves, and it must be smaller than the average number of the points within the same size of spherical search region (as listed in Table 6); otherwise, the filter will bring in over-filtering and the point cloud will break into numerous pieces. The value $k$ is the size of the point set which is used to calculate the principal component vector by PCA. Following the suggestion in [13], the value of $k$ is fixed to 20 for all point clouds processed in this paper. Parameter $d_1$ is a threshold on distance for the 3D region growing on the leaf center pre-segmentation stage. The pre-segmentation reaches optimal when $d_1$ is set about two to four times of the average spacing. If $d_1$ is set too large, different leaf point clouds will be aggregated into one piece of leaf. Conversely, a single leaf center may break into pieces if $d_1$ is set too small. The configuration of parameters in the facet over-segmentation stage can refer to suggestions in [4].

**Table 6.** Average spacing and the average number of points in a spherical neighborhood for the four point clouds.

|  |  | Epipremnum aureum | Monstera deliciosa | Calathea makoyana | Hedera nepalensis |
|---|---|---|---|---|---|
| Average spacing (meter) |  | 0.00199 | 0.00257 | 0.00138 | 0.00168 |
| The average number of points within radius $R$ ($R = a \times Average\_spacing$) | $a = 2$ | 8 | 9 | 7 | 8 |
|  | $a = 3$ | 20 | 23 | 16 | 20 |
|  | $a = 5$ | 55 | 63 | 43 | 60 |
|  | $a = 10$ | 190 | 230 | 164 | 235 |
|  | $a = 50$ | 2605 | 2653 | 3874 | 4722 |

**Table 7.** Actual values of parameters used for segmenting the four types of point clouds, respectively

| Parameter | Description | Epipremnum aureum | Monstera deliciosa | Calathea makoyana | Hedera nepalensis |
|---|---|---|---|---|---|
| $r$ | The radius parameter in the 3D joint filter. | 0.01m | 0.01m | 0.005m | 0.004m |
| $n_{threshold}$ | A threshold that defines the minimum number of points within a sphere of radius $r$. | 40 | 15 | 13 | 8 |
| $k$ | The number of neighboring points used in PCA. | 20 | 20 | 20 | 20 |
| $d_1$ | A threshold for region growing based pre-segmentation. | 0.004m | 0.006m | 0.004m | 0.006m |

## 7. Discussion

*7.1. Applying the 3D joint filtering for multiple times*

In this paper, the plant point cloud scanned from the binocular stereo vision platform and the Kinect V2 require the 3D joint filtering only once, while the point cloud from the multi-view platform needs three or four repeats because the point cloud acquired from multi-view imaging is denser than the others. Generally, for crowded point clouds, 3D joint filtering should be applied for multiple times. It is possible that after several times of joint filtering, only a few central areas of leaves remain or even the whole point cloud is filtered out. In the two cases, the proposed method can still achieve satisfactory segmentation results because the completely-filtered leaves will be labeled as new leaves during facet region growing. However, insufficient 3D joint filtering is not able to separate seriously overlapped leaves in a dense plant point cloud, resulting in under-segmentation. Thus, it is recommended to apply the 3D joint filtering operator for multiple times to filter all overlapping leaf areas.

Figure 16 demonstrates the comparison of leaf segmentation results for the *Calathea makoyana* point cloud under different times of 3D joint filtering. In Figure 16, from the first row to the fourth row are the leaf segmentation results with one, three, four, and five times of 3D joint filtering, respectively. The last three rows are satisfactory results, which not only correctly segment all 21 leaves in the point cloud, but also obtain high average cover rates (higher than 99%). It can be noted that after applying the 3D joint filtering operator for more than 3 times, there are very few leaf centers left, even nothing remains after 5 times of filtering. However, we can still segment all leaves perfectly by using the facet region growing algorithm proposed in Table 3 to grow current leaf labels and to discover new leaves in the filtered parts.

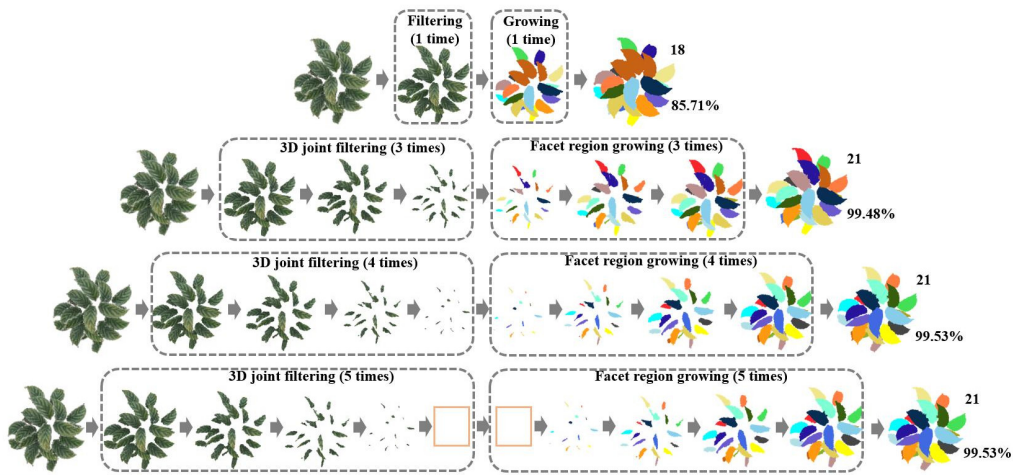

**Figure 16.** The comparison of leaf segmentation results for *Calathea makoyana* point cloud under different times of 3D joint filtering. From the first row to the fourth row are the leaf segmentation results with one, three, four, and five times of 3D joint filtering, respectively. The number on the upper-right side of each final segmentation represents the number of successfully segmented leaves, and the value on the lower-right corner is the average segmentation accuracy of all leaves.

*7.2. Application to estimation of leaf areas*

The proposed individual leaf segmentation method can help to estimate leaf traits such as area, length, and width in a fully-automatic way.

Due to the rugged surfaces of real leaves, it is difficult to directly measure the area of each single leaf in a canopy; however, we can first scan the canopy into a point cloud, and then apply leaf area calculation on the segmented point cloud. Therefore, we design a new method to estimate the leaf area based on the point cloud of each single leaf, and the steps are given as follows. Firstly, the leaf point cloud is down-sampled and smoothed to further reduce the influences from outliers on the leaf surface. Secondly, the leaf point cloud is turned into a mesh of a large number of greedy projection triangles by the method in [35]. Thirdly, the area of each spatial triangle is calculated through the coordinates of the three vertices, and the leaf area is calculated as the sum of all triangle areas on that leaf point cloud. Figure 17 demonstrates the leaf area estimation process and the ground truth generation. Figure 17(a) illustrates the triangulation result for a segmented individual leaf point cloud of *Calathea makoyana*; Figure 17(b) enlarges an small area in 17(a) to show the detail of generated triangles; and the area of $\triangle ABC$ in Figure 17(b) is calculated by $1/2 \cdot \mathbf{AB} \times \mathbf{AC}$. Figure 17(c) is a rendered model by meshing on 17(a), we believe that the surface area of the leaf model in 17(c) can be approached by the sum of all triangle areas in 17(a). The ground truth of each leaf area is calculated by an easy image processing technique. We clamp each leaf between two parallel pieces of glasses, and place them right below a camera to capture the image that contains the flattened leaf and a $16cm^2$ square reference paper as in Figure 17(d). Afterwards, in Figure 17(f) the ground truth of the leaf area can be computed by comparing the number of leaf pixels with the number of pixels in the reference area. The ground truth of leaf length and width can also be fixed by finding the bounding box of the leaf contour in Figure 17(f) that has the largest area. We compare the estimated leaf area with the ground truth for nine different leaves from the point cloud of *Calathea makoyana* in Figure 18. The leaf length and width are two orthogonal traits. The two values are estimated by projecting the leaf onto the plane with the average leaf normal and then searching the longest point distribution as the leaf length and its point distribution on the orthogonal direction as the width. The average error of our leaf area estimation technique is around 0.47%, and the average errors of the leaf length and width estimation are 2.89% and 4.64%, respectively. The results not only prove the high accuracy of our leaf traits estimation, but also reveals the effectiveness of the proposed individual leaf segmentation framework.

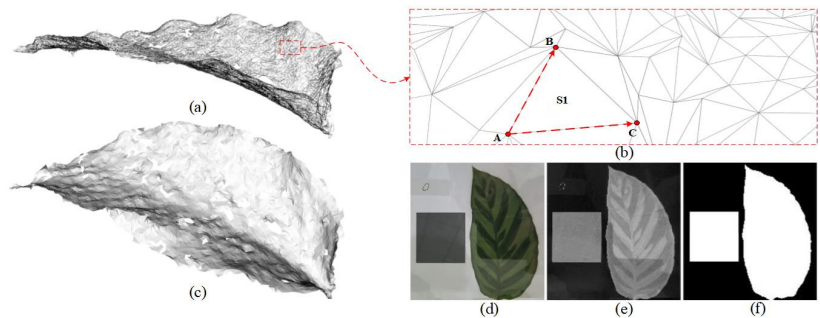

**Figure 17.** Illustrations of our leaf area estimation technique and the ground truth generation. (a) shows the triangulation result for a segmented individual leaf point cloud of *Calathea makoyana* after down-sampling and smoothing; (b) is a locally enlarged area from (a) for showing details of the generated triangles, in which the area of each triangle is calculated as to $\triangle ABC$ by $1/2 \cdot \mathbf{AB} \times \mathbf{AC}$. (c) is a rendered model of (a) viewed from another angle. The ground truth generation is shown by (d), (e), and (f). In (d) we clamp the leaf to be estimated between two parallel pieces of glasses, and place them right below a camera to capture the image that contains the flattened leaf and a $16cm^2$ square reference paper. Then we use grayscale processing to turn (d) into a grayscale image, and finally threshold the image of (e) into a binary image (f) that only contains the leaf foreground pixels and the reference pixels. By comparing the pixels belong to the leaf and the pixels belong to the reference paper, the leaf area can be estimated. The ground truth of leaf length and width can also be fixed by finding the bounding box of the leaf contour in (f) that has the largest area.

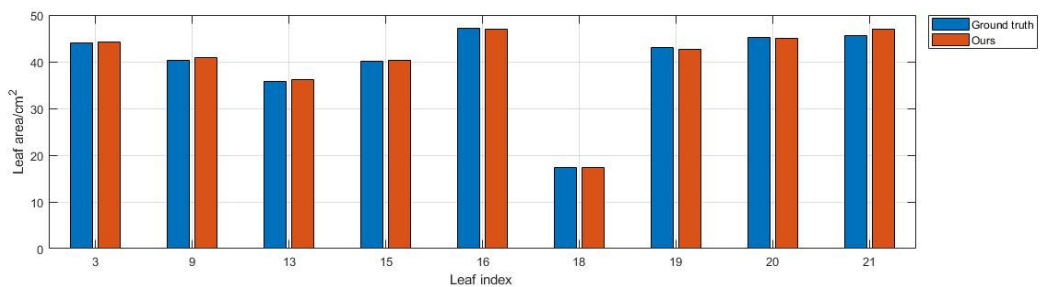

**Figure 18.** The comparison between the estimated leaf area and the ground truth for nine leaves in the point cloud of *Calathea makoyana*. The horizontal axis of this bar chart stands for the index of the leaf in the point cloud, and the vertical axis is the leaf area in $cm^2$. Bars in blue represent the ground truth of leaf areas obtained by the generation process shown in Figure 17. The bars in orange represent the leaf area estimation results of our technique. The estimated leaf area is very close to the ground truth, for all observed nine leaf samples.

## 8. Conclusions

In order to address the issue of inaccuracy in segmenting individual leaves from plant point clouds with serious leaf overlapping, we propose an overlapping-free individual leaf segmentation approach for plant point clouds by integrating 3D filtering and facet region growing. The method can be divided into three steps. Firstly, the occluded and overlapped parts among leaves in the point cloud are filtered out by a novel 3D joint filtering operator, and the remaining areas of leaf centers are pre-segmented into basic leaf sets. Secondly, the facet over-segmentation algorithm is employed on the filtered areas from previous 3D joint filtering to create a set of facets that separate the filtered points into clusters, and then the over-segmented facets are added back to pre-segmented leaf centers in the previous step. At last, we begin from the labeled leaf centers to search unlabeled adjacent facets and grow labels from inside to outside. The segmentation is complete when all facets are labeled. Experiments are carried out on four types of plant point clouds acquired from three different kinds of 3D imaging platforms. The results show that the proposed method is effective and efficient on segmenting individual leaves from crowed plant point clouds. In addition, we also present techniques for estimating leaf traits such as the area, length, and width.

Currently, the proposed method still has some restrictions. First, several parameters need to be tuned for an optimal segmentation result. So, we decide to add an adaptive mechanism for parameter tuning in the future. Second, the pre-processing step that concatenates several different filters is tailor-made for each species. Therefore, we also hope to utilize advanced approaches such as deep learning to carry out preprocessing for raw point clouds and to conduct pre-segmentation for leaf centers to lower the bar for implementation.

**Funding:** This work was supported in part by the Fundamental Research Funds for the Central Universities of China under Grants 2232019G-09, 2232017D-13, GSIF-DH-M-2019009, and also in part by the National Natural Science Foundation of China under Grants 61603089, 61603090.